\pdfoutput=1
\documentclass[11pt]{article}

\usepackage{acl}

\usepackage{times}
\usepackage{latexsym}

\usepackage{graphicx}               % Include graph
\usepackage{tabularx}               % Better table formatting
\usepackage{booktabs}
\usepackage{amsmath}
\usepackage{stmaryrd}
\usepackage{xcolor}
\usepackage{soul}

\newcolumntype{C}{>{\centering\arraybackslash}X}
\usepackage{multirow}               % Multi-row in tables
\usepackage{diagbox}                % Diagonal line in tables
\usepackage{hhline}                 % Draw double line in tables
\usepackage{color}                  % Text and background color
\usepackage{amsmath}                % Formula
\usepackage{amssymb}                % Formula
\usepackage{mathtools}              % Formula
\usepackage{enumitem}               % Better itemize environment

\usepackage{subfigure}
\usepackage{booktabs}
\usepackage{extarrows}
\usepackage{makecell}

\definecolor{RoseQuartzBg}{HTML}{F7CAC9}
\definecolor{RoseQuartz}{HTML}{F5A798}
\definecolor{Serenity}{HTML}{92A8D1}
\definecolor{OrangeRed}{rgb}{1.0, 0.27, 0.0}
\definecolor{Red}{rgb}{1.0, 0.0, 0.0}
\definecolor{Turquoise}{HTML}{0F4C81}
\usepackage{xparse}
\NewDocumentCommand{\lifu}{ mO{} }{\textcolor{OrangeRed}{\textsuperscript{\textit{Lifu}}\textsf{\textbf{\small[#1]}}}}
\NewDocumentCommand{\zhiyang}{ mO{} }{\textcolor{blue}{\textsuperscript{\textit{zhiyang}}\textsf{\textbf{\small[#1]}}}}
\NewDocumentCommand{\deval}{ mO{} }{\textcolor{Red}{\textsuperscript{\textit{Deval}}\textsf{\textbf{\small[#1]}}}}
\NewDocumentCommand{\apoorv}{ mO{} }{\textcolor{Red}{\textsuperscript{\textit{Apoorv}}\textsf{\textbf{\small[#1]}}}}
\usepackage[T1]{fontenc}
\usepackage[utf8]{inputenc}

\usepackage{microtype}

\usepackage{multirow}
\usepackage{booktabs}

\usepackage{xcolor,colortbl}
\definecolor{Gray}{gray}{0.95}

\usepackage[prependcaption,textsize=tiny]{todonotes}

\title{Probing Script Knowledge from Pre-Trained Models}

\author{Zijia Jin$^{\P}$, \  Xingyu Zhang$^{\natural}$, \ Mo Yu$^{\clubsuit}$, \ Lifu Huang$^{\spadesuit}$
\\
$^{\P}$New York University, \
$^{\natural}$Xi'an Jiaotong University, \
  $^{\clubsuit}$WeChat AI, \
  $^{\spadesuit}$Virginia Tech 
 \\
 $^{\P}${\tt zj2076@nyu.edu}, \
 $^{\natural}${\tt xy.zhang@stu.xjtu.edu.cn}, \\
  $^{\clubsuit}${\tt moyumyu@tencent.com}, \
  $^{\spadesuit}${\tt lifuh@vt.edu}
  }

\begin{document}
\maketitle

\begin{abstract}

Script knowledge is critical for humans to understand the broad daily tasks and routine activities in the world. Recently researchers have explored the large-scale pre-trained language models (PLMs) to perform various script related tasks, such as story generation, temporal ordering of event, future event prediction and so on. However, it's still not well studied in terms of how well the PLMs capture the script knowledge. To answer this question, we design three probing tasks: \textit{inclusive sub-event selection}, \textit{starting sub-event selection} and \textit{temporal ordering} to investigate the capabilities of PLMs with and without fine-tuning. The three probing tasks can be further used to automatically induce a script for each main event given all the possible sub-events. Taking BERT as a case study, by analyzing its performance on script induction as well as each individual probing task, we conclude that the stereotypical temporal knowledge among the sub-events is well captured in BERT, however the inclusive or starting sub-event knowledge is barely encoded.

% Script knowledge is one of the most complex stereotypical knowledge in human mind. Since it not only contains temporal relation between two events(\textit{Temporal Knowledge}), but also know about which events are suitable for specific scenes(\textit{Inclusive  Knowledge}). Recently, more researchers begin to use pre-trained language models(PLMs) to improve model's performance in Script Knowledge tasks. So it is crucial to answer these problems. What kind of Script Knowledge are already encoded during  pretraining process? What is the best way to induce this knowledge from PLMs? How to evaluate Script Knowledge via downstream task? To answer all these questions, we firstly design different tasks to probe PLMs performance in \textit{Inclusive Knowledge} and \textit{Temporal Knowledge}. And then we combine them together to generate the whole scripts and evaluate scripts quality with Rouge-L metric. Finally, we also implement some prompt methods and find that Bert captures enough Script Knowledge from pretraining but need a good format to induce them.
\end{abstract}
\section{Introduction}

A script is a structure that describes a stereotyped sequence of events that happen in a particular scenario~\cite{schank1975scripts,schank2013scripts}. It allows human to keep track of the states and procedures that are necessary to complete various tasks from daily lives to scientific processes. Taking the task of \emph{Eating in a Restaurant} as an example. A classic example script for this task may consist of a chain of subevents, such as \emph{Enter}$\rightarrow$\emph{Order}$\rightarrow$\emph{Eat}$\rightarrow$\emph{Pay (and Tip)}$\rightarrow$\emph{Leave}. The script knowledge has shown benefit to many downstream applications, such as story generation~\cite{li2013story,li2018constructing,guan2019story,zhai2019hybrid,lin2022inferring}, machine reading comprehension~\cite{tian2020scene,ostermann2018mcscript,sugawara2018makes}, commonsense reasoning~\cite{ding2019event,huang2019cosmos,bauer2021identify} and so on.

Recent large-scale pre-trained language models (PLMs)~\cite{devlin2019bert,liu2019roberta,radford2019language,raffel2019exploring} have shown competitive performance on many natural language processing tasks. Abundant studies have demonstrated that these models either directly capture certain types of syntactic~\cite{goldberg2019assessing,clark2019does,htut2019attention,rosa2019inducing}, factual~\cite{petroni2019language,petroni2020context,bouraoui2020inducing,wang2020language} and commonsense knowledge~\cite{zhou2020evaluating,rajani2019explain,lin2020birds} during the pre-training or acquire inductive capability to more efficiently induce such knowledge from natural language text~\cite{pandit2021probing,bosselut2019comet}. However, as another important type of cognitive and schematic knowledge describing human routine activities, scripts are not yet well probed in the language models by prior studies.  

To investigate how well the pre-trained language models have captured the script knowledge, in this work, we design three probing tasks and language model prompting methods to probe the script knowledge from PLMs, and further leverage the language model prompting methods to induce the scripts given the main events. Specifically, we aim to answer the following two research questions:

\textbf{\emph{Whether and what script knowledge is captured by the pre-trained language models.}} To answer this question, we design three sub-tasks to probe the script knowledge, including \textbf{inclusive sub-event selection} (i.e., whether a sub-event is included or excluded in a main event or task), \textbf{starting sub-event selection} (i.e., which sub-event is the start of the script for a particular main event), and \textbf{sub-event temporal ordering} (i.e., predicting a temporal before or after relation between two sub-events). On these sub-tasks, we explore both template-based and soft prompting methods to query the knowledge from pre-trained language models. By investigating their performance gaps to the fine-tuning results, we find that both the inclusive and starting sub-event selection sub-tasks have relatively poorer performance than that of temporal ordering, which is likely due to the lack of relevant objectives to encourage the models to capture such knowledge during pre-training, and further suggests future research directions to enhance the PLMs to better capture the script knowledge.
%indicating the lack of the sub-event relation knowledge in the pre-trained models.

\textbf{\emph{How to better generate the scripts from these pre-trained models.}} With the language model prompting methods, we can select the inclusive sub-events of a particular script, the starting sub-event and subsequent events by predicting the temporal order among all the inclusive sub-events, which can ultimately generate a sequence of events as the script of a main event. Thus, we further design a benchmark dataset to fine-tune the models for the three sub-tasks and evaluate their performance on generating the whole scripts for various main events from diverse domains and topics.

% Though the two previously defined sub-tasks, i.e., sub-event selection and ordering, can be well-solved by the pre-trained language models, they are not sufficient to accurately generate the whole event sequences which is the ultimate goal of script generation. Thus, we further propose an auto-regressive approach with two new sub-tasks including \textbf{starting sub-event prediction} (i.e., predict the starting sub-event from all selected sub-events for the main event or task), and \textbf{subsequent sub-event prediction} (i.e., predict the subsequent sub-event given the target task and preceding events), and accordingly design two prompting methods to tackle these tasks. Experiments show that our approach extracts whole scripts with advanced performance.

The contributions of this work can be summarized as follows:
\vspace{-5pt}
\begin{itemize}[leftmargin=*]
\setlength\itemsep{0em}
\item We are the first to formulate the sub-tasks and set up benchmark datasets to probe the script knowledge from pre-trained language models.
\item We are the first to research on the generation and evaluation of the whole scripts from pre-trained language models.
\end{itemize}

\section{Related Work}
\paragraph{Script Knowledge}
The definition of Script Knowledge was first proposed in 1981~\cite{feigenbaum1981handbook}, which aims to detect the relation between two events.~\newcite{DBLP:conf/acl/ChambersJ08} created the first unsupervised data-driven method based on point-wise mutual information (PMI) to automatically extract narrative event chains. Recently, researchers explored deep neural networks, especially large-scale pre-train language models to predict the temporal relation between two events~\cite{pustejovsky2003timebank,DBLP:conf/emnlp/Chambers13, DBLP:conf/aaai/FerraroD16, DBLP:conf/acl/ReimersDG16} or generate the future event~\cite{DBLP:conf/eacl/PichottaM14,DBLP:conf/eacl/JansBVM12,DBLP:conf/emnlp/ZhangLC20}. Comparing with these studies, our work focuses more on investigating how well the PLMs encode or capture the script knowledge from pre-training and their bottleneck, suggesting possible directions for future research.
%more and more people turned to use language models to predict the relation between two events\cite{DBLP:conf/emnlp/Chambers13, DBLP:conf/aaai/FerraroD16}. Some people also begin to use PlMs to predict or generate next event\cite{DBLP:conf/emnlp/ZhangLC20}. The main difference with our work is that they are trying to utilize PLMs to answer these problems better but we are trying to see what kind of knowledge PLMs already learned.

\paragraph{Language Model Probing}
Probing is a popular way to detect what knowledge is encoded in PLMs. At first, probing method is designed for detect morphology knowledge\cite{belinkov-etal-2017-neural} ,syntactic knowledge \cite{peters-etal-2018-deep} and semantic knowledge\cite{DBLP:conf/iclr/TenneyXCWPMKDBD19}. Then researchers began to pay more attention to more complex knowledge like commonsense knowledge. The two main standard approaches in probing commonsense knowledge is building classifiers\cite{DBLP:conf/emnlp/HewittL19} or filling text in the gap\cite{petroni-etal-2019-language}. In our study, we extend the accuracy based methods and designed a series of downstream tasks specific to Scripts Knowledge.

% \paragraph{Related Resources for Script Knowledge Learning or Induction of the Script FSM}
% Finally, there are works DeScript~\citep{DBLP:conf/lrec/WanzareZTP16} and OMICS~\citep{DBLP:conf/aaai/GuptaK04}.
% Their datasets provide instances of scripts in natural language texts. 
% These works have a different goal from ours, which is usually to extract a graph of event transitions for a scenario (e.g., a finite state machine) that can generate valid scripts of event sequences with correct temporal orders.
% While we have a different goal of probing the scripts from pre-trained models, we find their script instances a suitable resource for our evaluation.
\section{Method}
\begin{figure*}[!htpb]
    \centering
    \vspace{-6mm}
    \scalebox{0.28}{
    \includegraphics{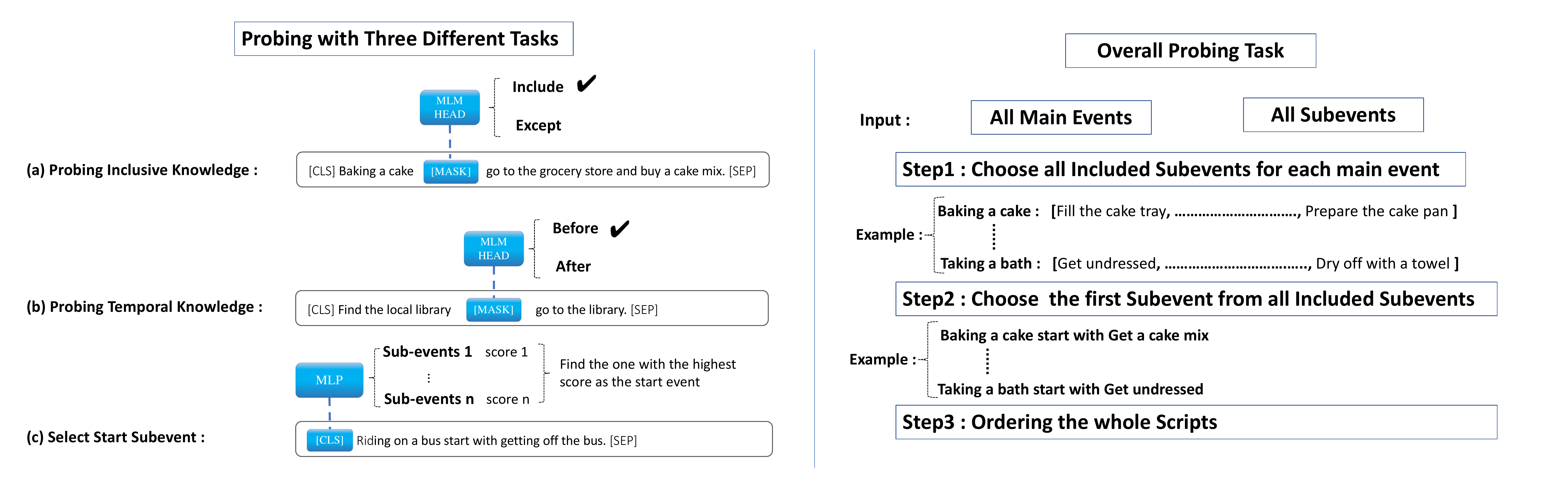}}
    \vspace{-8mm}
    \caption{Overview of the probing approaches for (1) Inclusive Sub-event Selection, (2) Starting Sub-event Selection, and (3) Sub-event Temporal Ordering. And an overall evaluation stage for generating scripts with main events and subevents as input.}
    \label{fig:pipeline}
    \vspace{-4mm}
\end{figure*}
%\lifu{need to add an overview Figure to show the four prompting methods for the four tasks}

% In this section, we first separately introduce the SSR-Local-Probe framework and SSR-Global-Probe framework and use it to detect current  PLMs'  ability to encode scripts knowledge. Then, we use some automatically prompt methods like P-tuning to improve the models' performance on those probing tasks.

% \subsection{SSR Local Probe}
% In order to comprehensive assess PLMs models' capabilities in scripts learning, we divide the whole process into three local evaluate tasks, including a \textit{Local Subevent Selection} task, a \textit{Local First Subevent Selection} task and a \textit{ Local Subevent Ordering} task. 

\subsection{Script Knowledge Probing}
Our first goal is to probe the script knowledge from pre-trained language models. To do so, we divide the script knowledge into three categories: the \textit{Inclusive} and \textit{starting} relation between each sub-event and main event, indicating whether the sub-event should be included in or the start of the script of a particular main event, and the \textit{temporal} relation (i.e., \textit{Before} or \textit{After}) among the sub-events. To probe these knowledge from PLMs, we design the following tasks.
%(1) the \textit{Inclusive} or \textit{Exclusive} relation between each sub-event and main event; (2) the \textit{Starting} subevent knowledge which indicates the start of the script for each main event; and (3) the temporal knowledge (i.e., \textit{Before} or \textit{After}) among subevents from the same script. To probe these two categories of script knowledge from PLMs, we design two following tasks.

\paragraph{Task 1: Inclusive Sub-event Selection} 
As Figure~\ref{fig:pipeline} shows, given a main event, e.g., \textit{"Clean laundry"}, and a candidate sub-event,  e.g., \textit{"Gather dirty clothes."}, we aim to have the language model to determine whether the sub-event belongs to the script of the target main event. To do so, we use \texttt{[MASK]} to connect them into a whole sequence and use a PLM to encode the sequence into contextual representations. In order to predict the \textit{Inclusive} relation, we apply a linear function (i.e., a MLM head) to project the \texttt{[MASK]} into a probability distribution over the whole vocabulary of the PLM. By exploring many candidate tokens from the target vocabulary to represent each relation, we finally select ``include'' to denote the \textit{Inclusive} relation and ``except'' for \textit{Exclusive}.

%and predict a binary relation (i.e., \textit{Include} or \textit{Exclude}) based on the contextual representation of \texttt{[MASK]}. By exploring many candidate tokens from the target vocabulary to represent each relation, we finally select ``include'' to denote the \textit{Include} relation and ``except'' for \textit{Exclude}.

%a language model prompting approach is designed to induce the inclusive knowledge from PLMs, that is, selecting a subset of subevents belonging to the script of the target main event. 
% As Figure~\ref{fig:pipeline} shows, given a main event, e.g., \textit{EXAMPLE OF MAIN EVENT}, and a candidate sub-event, e.g., \textit{EXAMPLE OF SUB-EVENT}, we use \texttt{[MASK]} to connect them into a whole sequence. Then we use a PLM to encode the sequence and predict a binary relation (i.e., \textit{Include} or \textit{Exclude}) based on the contextual representation of \texttt{[MASK]}. By exploring many candidate tokens from the target vocabulary to represent each relation, we finally select ``include'' to denote the \textit{Include} relation and ``except'' for \textit{Exclude}.

\paragraph{Task 2: Starting Sub-event Selection} Given a main event and a set of sub-events that are predicted to belong to the script of the main event, we aim to select the most probable sub-event as the start of the script. We formulate it as a  sequence classification problem. We concatenate the main event and each sub-event candidate with a prompt "\textit{start with}", e.g., \textit{Taking bus start with finding bus stop}, and use a MLP layer to predict a score indicating how likely the sub-event is the start of the script of the main event, based on the contextual representation of the \texttt{[CLS]}. As a result, we use the sub-event with the highest score as the first sub-event. We design a margin based loss function to encourage the score of the positive start sub-event to be higher than others
%we compare the logits belongs to True label and find the subevent with the largest logit as the start event. %. During training process, we encode these sentences with PLMs and use a MLP layer to predict the score of each subevent, and predict the subevent with highest score as the positive sample. 
\begin{equation*}
    L(s^{*}, s_i) = \sum_{\tilde{s}_i\in{\tilde{S}}}max(score(\tilde{s}_i) + m - score(s^{*}), 0)
\end{equation*}
where $s^{*}$ represents the positive start sub-event of a particular script and $\tilde{S}$ denotes the set of other sub-events from the same script. The margin $m$ is a hyper-parameter, which is set as 1.0 in our experiment. During inference, given a set of candidate sub-events, we compare their scores and select the one with the highest score as the starting sub-event. 

% \lifu{confusing: are you predicting a score or a binary label? if binary label, is there a loss function for this binary classification? Answer: we do not make a classification, we recognize the subevent with the highest score as the first one.}
% \lifu{If you are not predicting a binary label, re-draw the figure}

%Meanwhile, in order to improve the model performance we use a designed margin loss function shows in Equation~\ref{margin} to make the positive data get higher score than others. While inference, we compare the logits belongs to True label and find the subevent with the largest logit as the start event.

\paragraph{Task 3: Sub-event Temporal Ordering} This probing task is to show the capability of the PLMs on correctly organizing the sub-events into a temporally ordered event sequence. To do so, we design a new language model probing approach following~\cite{DBLP:conf/emnlp/PetroniRRLBWM19}. As shown in Figure~\ref{fig:pipeline}, given two subevents, e.g., \textit{"put clothes in dryer."} and \textit{"turn on dryer."} , we use \texttt{[MASK]} to connect them into a sequence and use a PLM to encode it. The temporal relation is predicted by comparing the probability of tokens ``\textit{before}'' and ``\textit{after}'' based on the contextual representation of \texttt{[MASK]}.

\subsection{Script Induction with PLMs}

The second goal in this work is to design a simple yet effective approach to automatically induce scripts based on PLMs. Given a particular main event and a set of candidate sub-events, to induce the script for the target main event, we design a pipeline approach consisting of three steps: (1) selecting a subset of inclusive sub-events from all the candidates; (2) determining the starting sub-event; and (3) ordering all the inclusive sub-events by predicting the temporal relation between each pair of them. These three steps correspond to the three approaches designed for script knowledge probing.

\section{Experiment Setup}

We take BERT-base-uncased~\cite{devlin2019bert} as the target PLM to investigate how well it encodes the script language via the three probing tasks. We combine three script datasets, including DeScript~\cite{DBLP:conf/lrec/WanzareZTP16}, OMICS~\cite{DBLP:conf/aaai/GuptaK04} and Stories~\cite{DBLP:journals/corr/abs-1806-02847}, where each main event is annotated with 7 to 122 scripts written by different crowd-sourcing workers. We sample 60 main events as the evaluation set, 39 main events as the development set and use the remaining 98 main events for training. For the main events in training and development sets, we keep all the scripts, while for each main event in the evaluation set, we only keep the longest script as the target. Table~\ref{tab:task_statistics} shows the statistics of each dataset.

%\lifu{describe how you create the candidate event pool, why you keep only the longest script for evaluation set and keep all for training} 

%We select three different datasets as our Scripts Knowledge base, including, DeScript, OMICS and Stories. The details of datasets statics are shown in \ref{tab:data_statistics}. And we combine them together and spilt them into two part, one is data for training in different tasks we call it local data, the other is data for overall evaluation we call it global data.

%In this paper, we first probe BERT models' abilities in capturing Scripts Knowledge via different sub-tasks including \textit{Inclusive Subevent Selection Task}, \textit{Temporal Order Task} and \textit{Starting Subevent Selection Task}. We use those three tasks to evaluate what kind of Scripts Knowledge does BERT encodes via pretraining  process? So we first train those three tasks in different datasets. And then we combine those separate tasks and models together and use them to evaluate BERT's ability in \textit{Overall Evaluation} (Overall means giving main events and many subevents, use models to generate a whole script for each main event ) with test data.

% We select three different datasets as our Scripts Knowledge base, including, DeScript, OMICS and Stories. The details of datasets statics are shown in \ref{tab:data_statistics}. And we combine them together and spilt them into two part, one is data for training in different tasks we call it local data, the other is data for overall evaluation we call it global data.
\begin{table}[ht]
	%\vspace{-2mm}
% 	\resizebox{1\columnwidth}{!}{
		\begin{tabular}{l|c|c}
			\hline
			\toprule
			\textbf{Datasets} & \textbf{\# Main Events} & \textbf{\# Scripts}\\
			\hline
			\textbf{Training} & 98 &  4,685 \\
			\textbf{Development} & 39 &  1,791 \\
			\textbf{Evaluation} & 60 & 60 \\
			\bottomrule
			\hline
		\end{tabular}
		\caption{Data statistics for training, development and evaluation Sets.}
	\label{tab:task_statistics}
		%\vspace{-2.mm}
\end{table}

To create the training samples for the \textit{inclusive sub-event selection} task, for each script, we use all the ground truth subevents as positive samples and randomly choose 100 times of negative samples from other main events' scripts. For evaluation, as the inclusive sub-event selection requires a pool of all the possible candidate events, we combine the sub-events of all scripts in the evaluation dataset. To create the training samples for the \textit{start sub-event selection} task, we use the first sub-event of each script as the positive sample and all the remaining sub-events from the same script as the negative samples. During the inference, we select the starting sub-event from the inclusive sub-events predicted by the inclusive sub-event selection approach. We use accuracy as the evaluation metric. Finally, for the temporal ordering task, we create each training sample based on each sub-event together with one of its following sub-events. We randomly shuffle the order of each pair of sub-events and create its corresponding label: \textit{"before"} or \textit{"after"}. To evaluate the quality of the temporal ordering among all the sub-events, we first generate a script based on the predicted temporal order and then use ROUGE-L to evaluate the longest common subsequence between the generated script and the gold script. 

% all scripts as input and mark the first subevent as the start event. During the inference progress, we use models to select start events from all ground truth inclusive results and use accuracy as the metric. For temporal ordering task, we build the training samples by combining subevents and its next subevents with \texttt{[MASK]} tokens in each scripts and make the model to predict it is \textit{"before"} or \textit{"after"}. However, we only predict which one is the next with highest score during evaluation. 

We compare the following approaches for each probing task as well as the script induction:

\paragraph{BERT Pre-trained:} Directly use the pre-trained BERT model to make the predictions on the evaluation set.

\paragraph{BERT Fine-tuning:} Fine-tune BERT with task-specific training data and evaluate those fine-tuned models on the evaluation set. 

\paragraph{BERT Ptuning:} Following the Ptuning framework~\cite{liu2021gpt}, fine-tune the parameters of both BERT model and prompt tokens.

\paragraph{BERT Ptuning Freeze:} Only fine-tune the prompt tokens while freezing the parameters of BERT model.

\section{Results and Analysis}

\subsection{Overall Script Induction}
\label{sec:overall}
We first show the results of end-to-end script induction given each main event and the pool of all candidate sub-events. As Table~\ref{tab:overall_res} shows, without any fine-tuning, BERT-Pretrained can barely induce any reasonable scripts. The high precision and low recall indicates that the bottleneck is likely in correctly selecting the inclusive sub-events for each main event. However, with fine-tuning either on the whole BERT parameters or a few prompt parameters, the script induction performance can be improved significant, demonstrating that the pre-trained BERT actually captures certain level of script knowledge but requires external probes to induce such knowledge from it. Finally, by analyzing of the performance of fine-tuning approaches, we notice a more significant improvement on recall. We conjecture that with fine-tuning, the inclusive sub-event selection is more likely to be improved.

%We first show the overall evaluation results in table~\ref{overall}. Firstly, zero-Shot BERT results shows that it can only capture little scripts knowledge during pre-training process. Secondly, we see that our fine-tuning efforts do help improve model performance and gain 23.92 pt increment. Thirdly, we show the prompt method results in line 3. Under this setting, we only tuning the parameters in prompt tokens and fix all BERT parameters. The prompt method outperform the best results of BERT-based-finetuning model which shows that prompt words are more important in exporting script knowledge than finetuning. 

\begin{table}[h]
	\centering
	\resizebox{\columnwidth}{!}{
		\begin{tabular}{l|ccc}
			\hline
			\toprule
			\multirow{2}{*}{\textbf{Method}} & \multicolumn{3}{c}{\textbf{Rouge-L}} \\
			& \bf Rec & \bf Prec  & \bf F-score  \\
			\midrule
			\textbf{BERT-Pretrained}  & 3.25 & 22.60 & 4.81  \\
			\textbf{BERT-Finetuning}  & 37.19 & 28.07 & 28.73  \\
			\textbf{BERT-Ptuning} & 48.70 & 28.78 & 32.52  \\
			\textbf{BERT-Ptuning-Freeze} & 4.82 & 16.19 & 7.25  \\
			\bottomrule
			\hline
		\end{tabular}}
	\caption{Performance of script induction}
	\label{tab:overall_res}
\end{table}

%\paragraph{Does BERT already have those Knowledge without Finetuning?} Compared the ptuning method result and zero-shot result, we can find that although we finetune prompt tokens it can not get better results than zero-shot BERT. However, It shows that BERT with finetuning prefix is more likely to improve the recall score and much lower precision score. It seems like that Bert-base do not contain much script knowledge during pretraining process.

\begin{table*}[ht]
	\vspace{0mm}
	\centering
	\resizebox{2\columnwidth}{!}{
		\begin{tabular}{l|ccc|c|c}
			\hline
			\toprule
			\multirow{2}{*}{\textbf{Method}} & \multicolumn{3}{c|}{\textbf{Inclusive Subevent Selection}}  & \multicolumn{1}{c|}{\textbf{Starting Subevent Selection}} & \multicolumn{1}{c}{\textbf{Temporal Ordering}} \\
			& \bf Rec & \bf Prec & \bf F-score  & \bf Accuracy  &  \bf Rouge-L F1 \\
			\midrule
			\textbf{BERT-Pretrained}  & 7.44 & 0.64 &1.17 & 18.33  & 63.79\\
			\textbf{BERT-Finetuning}  & 33.83 & 44.71 &38.51 & 21.66  & 62.87  \\
			\textbf{BERT-Ptuning} & 31.16 & 56.24 & 40.10 & 20.00  & 63.62  \\
			\textbf{BERT-Ptuning-Freeze} & 98.69 & 0.52 & 1.03 & 28.33  & 66.02  \\
			\bottomrule
			\hline
		\end{tabular}}
		\vspace{-2.mm}
		\caption{Performance on each individual task.}
	\label{3tasks}
\end{table*}

% \subsection{Overall Probing Results}
% We first show the overall evaluation results in table~\ref{overall}. Firstly, zero-Shot BERT results shows that it can only capture little scripts knowledge during pre-training process. Secondly, we see that our fine-tuning efforts do help improve model performance and gain 23.92 pt increment. Thirdly, we show the prompt method results in line 3. Under this setting, we only tuning the parameters in prompt tokens and fix all BERT parameters. The prompt method outperform the best results of BERT-based-finetuning model which shows that prompt words are more important in exporting script knowledge than finetuning. 

% \paragraph{Does BERT already have those Knowledge without Finetuning?} Compared the ptuning method result and zero-shot result, we can find that although we finetune prompt tokens it can not get better results than zero-shot BERT. However, It shows that BERT with finetuning prefix is more likely to improve the recall score and much lower precision score. It seems like that Bert-base do not contain much script knowledge during pretraining process.

\subsection{Probing on Individual Tasks}

We further analyze the capability of BERT on encoding each type of script knowledge based on the three probing tasks. To avoid error propagation, for both starting sub-event selection and temporal ordering, we use the gold inclusive sub-events of each main input as input.

As Table~\ref{3tasks} shows, for inclusive sub-event selection, without fine-tuning, both BERT-Pretrained and BERT-Ptuning-Freeze cannot correctly select any inclusive sub-events. This is likely due to the discrepancy between the pre-training objectives of BERT (i.e., MASK language modeling and next sentence prediction) with the objective of inclusive sub-event selection. With fine-tuning, the performance of both BERT-Finetuning and BERT-Ptuning is improved significantly, which is aligned with our assumption in Section~\ref{sec:overall}. Starting sub-event selection is hard to all the approaches, which is likely due to two reasons: one is the limited training samples, and the other is that though we formulate each sub-task as mask prediction to better induce the knowledge from BERT, the pattern ``\textit{Main\_Event} \textit{starts with} \textit{Sub\_Event}'' is less likely to appear in the unlabeled corpus than other patterns, such as ``\textit{Main\_Event} \textit{includes} \textit{Sub\_Event}'' and ``\textit{Event\_A} \textit{before/after} \textit{Event\_B}''. Finally, all the approaches show consistently descent performance on temporal ordering, no matter whether BERT is fine-tuned or not, demonstrating that BERT has well captured the relations among the events with
stereotypical temporal orders, possibly due to the next sentence prediction objective during pre-training.

\section{Conclusion}

In this work, we investigate the capability of large-scale pre-trained language models (PLMs) on capturing three aspects of script knowledge: \textit{inclusive sub-event knowledge}, \textit{starting sub-event knowledge} and \textit{temporal knowledge} among the sub-events from the same script. These three types of knowledge can be further leveraged to automatically induce a script for each main event given all the possible sub-events. We use BERT as a target PLM. By analyzing its performance on script induction as well as each individual probing task, we achieve the conclusions that the stereotypical temporal knowledge among the sub-events is well captured in BERT, however the inclusive and starting sub-event knowledge are not well encoded. 

% Script knowledge is critical for humans to understand the broad daily tasks and routine activities in the world. Recently researchers have explored the large-scale pre-trained language models (PLMs) to perform various script related tasks, such as story generation, temporal ordering of event, future event prediction and so on. However, it's still not well studied in terms of how well the PLMs capture the script knowledge. To answer this question, we design three probing tasks: \textit{inclusive sub-event selection}, \textit{starting sub-event selection} and \textit{temporal ordering} to investigate the capabilities of PLMs with and without fine-tuning. The three probing tasks can be further used to automatically induce a script for each main event given all the possible sub-events. Taking BERT as a case study, by analyzing its performance on script induction as well as each individual probing task, we conclude that the stereotypical temporal knowledge among the sub-events is well captured in BERT, however the inclusive or starting sub-event knowledge is barely encoded. 

% \section*{Acknowledgements}
% This document has been adapted

\bibliography{anthology,custom}
\bibliographystyle{acl_natbib}

\appendix

\section{Appendix}
\label{sec:appendix}

\end{document}